\documentclass[10pt]{cai}

\graphicspath{{figs/}}

\begin{document}

\volumeheader{35}{0}
\begin{center}

  \title{Multi-Scale Identity-Preserving Image-to-Image Translation Network for Low-Resolution Face Recognition}
  \maketitle

  \thispagestyle{empty}

  \begin{tabular}{cc}
    Vahid Reza Khazaie\upstairs{\affilone}, Nicky Bayat\upstairs{\affilone}, Yalda Mohsenzadeh\upstairs{\affilone,*}
   \\[0.25ex]
   {\small \upstairs{\affilone} University of Western Ontario} \\
  \end{tabular}
  
  \emails{
    \upstairs{*}ymohsenz@uwo.ca
    }
  \vspace*{0.2in}
\end{center}

\begin{abstract}

State-of-the-art deep neural network models have reached near perfect face recognition accuracy rates on controlled high-resolution face images (especially Caucasian faces). However, their performance is drastically degraded when they are tested with very low-resolution face images. This is particularly critical in surveillance systems, where a low-resolution probe image is to be matched with high-resolution gallery images. Super-resolution techniques aim at producing high-resolution face images from low-resolution counterparts. While they are capable of reconstructing images that are visually appealing, the identity-related information is not preserved. Here, we propose an identity-preserving end-to-end image-to-image translation deep neural network which is capable of super-resolving very low-resolution faces to their high-resolution counterparts while preserving identity-related information. We achieved this by training a very deep convolutional encoder-decoder network with a symmetric contracting path between corresponding layers. This network was trained with a combination of a reconstruction and an identity-preserving loss, on multi-scale low-resolution conditions. Extensive quantitative evaluations of our proposed model demonstrated that it outperforms competing super-resolution and low-resolution face recognition methods on natural and artificial low-resolution face data sets and even unseen identities.\\
\end{abstract}

\begin{keywords}{Keywords:}
Low-Resolution Face Recognition, Identity-Preserving Loss, Multi-Scale Super-Resolution, Image-to-Image Translation
\end{keywords}
\copyrightnotice

\section{Introduction}

Recent deep neural network models have demonstrated outstanding face recognition performance on various face data sets with challenges like variations in pose, illumination and occlusion and surpassed the performance of humans in these tasks \cite{He2019}. For example, most deep metric learning based models have accomplished over 99\% accuracy on the commonly used Labeled Faces in the Wild (LFW) \cite{huang2008labeled} benchmark. Despite that, their performance on some real-world applications where the faces are at very low-resolution like surveillance cameras is declined. The difference is most significant when the low-resolution face is below 32x32 pixels \cite{Chen2020}. A face with lower than 16x16 resolution can be even challenging to recognize with the human eye. Therefore, finding the most similar high-resolution gallery image for very low-resolution probe faces is still a challenging computer vision task which is investigated under low-resolution face recognition (LRFR) domain.

There are several methods to increase the performance of the models in low-resolution face recognition tasks \cite{zangeneh2020low}. One common solution is to super resolve the low-resolution faces to their high-resolution counterparts. However, most super-resolution methods are able to reconstruct faces with only rich visual features and the upsampled faces usually lack identity-related information \cite{Ledig2017} \cite{Dong2016}. So, those super-resolution models cannot enhance face recognition performance, which leads to substandard performance in face recognition. Recent studies proposed super-resolution methods that maintain identity information while reconstructing higher resolution images \cite{Chen2020} \cite{Hsu2019}. In line with this idea, we proposed a multi-scale identity-preserving deep convolutional encoder-decoder neural network which is capable of super resolving faces with richer semantic information and simultaneously enhancing the identity information optimized for face recognition. We attained this outcome by training the identity-preserving deep network with joint supervision. This supervision was a combination of the distances between faces in the pixel domain and in an identity-aware feature space produced by a pre-trained face recognition model. The main contributions of this work include:

\begin{itemize}
  \item We trained a very deep convolutional encoder-decoder neural network with skip-connections with an \emph{Identity-Preserved loss function} which is able to super-resolve very low-resolution faces while maintaining face identity information.
  \item Our model consists of contracting paths between encoder and decoder layers to capture context which enables it to localize the equivalent pixels more precisely. This increased the recognition performance on natural and artificial low-resolution faces.
  \item We designed a \emph{multi-scale low-resolution setup} to make our proposed model robust to various low-resolution conditions. In this setting, we trained our Identity-Preserving model on multiple different low-resolutions.
\end{itemize}


\section{Related Works}
\label{sec:relatedworks}

The low-resolution face recognition problem (LRFR) is a challenging subset of general face recognition tasks. One application of this problem is where the faces are captured from far distances like surveillance cameras. Most face recognition methods have nearly perfect performance on high-resolution faces but they perform poorly on low-resolution ones \cite{He2019}. The reason for this drop in performance is that the rich features required for face recognition cannot be extracted from tiny faces \cite{Yu2017}.
To solve this challenging problem, several approaches have been proposed, including super-resolution techniques. By super resolving low-resolution faces, their high-resolution peers which could be more suited to face recognition are reconstructed. However, most super-resolution models can reconstruct faces which are only visually pleasant and have insufficient identity-related information \cite{He2019}. 

Zhang et al. proposed a method using deep canonical correlation analysis (DCCA) for low-resolution face recognition and reconstruction \cite{zhang2018low}. Li et al. suggested a five-branch network based on human face key parts to generate high-resolution face key parts \cite{li2018face}. Zhu et al. used a  teacher-student learning setup for optimizing the features from a pre-trained model on high-resolution faces with a recognition and distillation loss \cite{zhu2019low}. Moreover, a suggested method by Noor et al. \cite{noor2019gradient} was to develop a residual neural network for super-resolution that enhances gradient images. Most recently, Yu and Porikli  \cite{Yu2020} utilized a discriminative network to embed spatial transformation layers. This allows local receptive fields to line-up with similar spatial supports and yields a better mapping between low-resolution and high-resolution facial patterns with a class-specific loss. While state-of-the-art face hallucination methods reconstruct visually appealing faces, they do not preserve structural identity information embedded in the low-resolution face. \cite{lai2019low} proposes a new loss function that combines with the image-content loss to supervise CNNs simultaneously for face hallucination and recognition. In order to overcome this problem, in the next section we proposed a very deep convolutional model with an identity-aware face super-resolution technique. These approaches try to increase resolution while protecting identity-related features of the image by adding an identity-aware loss term to the overall objective function when training the face hallucination system. Adding this feature to super-resolution models might compromise visual quality for better face recognition performance \cite{He2019} \cite{Hsu2019}.

\section{Method}
\subsection{Model Architecture}
\label{sssec:modelarchitecture}
Our proposed method includes two deep neural networks which are designed to receive low-resolution faces and produce the super-resolution counterparts which are optimized for both visual quality and face recognition. The first network is a deep convolutional encoder-decoder architecture with symmetric skip-connections between the corresponding layers (UNet-based architecture \cite{Ronneberger2015}) which helps to map the low-resolution and high-resolution face pairs more accurately through learning richer semantic representations and sharing low-level information across the network and the other network is a pre-trained face recognition model with an Inception-ResNet  architecture \cite{Schroff2015}.
Our identity-preserveing encoder-decoder network consists of 7 downsampling and 7 upsampling blocks with symmetric contracting paths (skip connections) for better localization of pixels between low-resolution and high-resolution pairs. In each downsampling block, we applied strided convolution with stride 2 followed by batch normalization and leaky relu layers. In each upsampling block, we used transposed convolution for doubling the spatial dimensions and also batch normalization, dropout, and relu activation functions. We added skip connections between each layer \emph{i} in the encoder and layer \( n-i+1 \) in the decoder, where \emph{n} is the total number of layers. Each skip connection simply concatenates all channels at layer \emph{i} with those at layer \( n-i+1 \). These skip-connections helped to localize the feature maps better through sharing information between the encoder and the decoder.

\begin{figure*}[t]
\centerline{\includegraphics[width=1\linewidth]{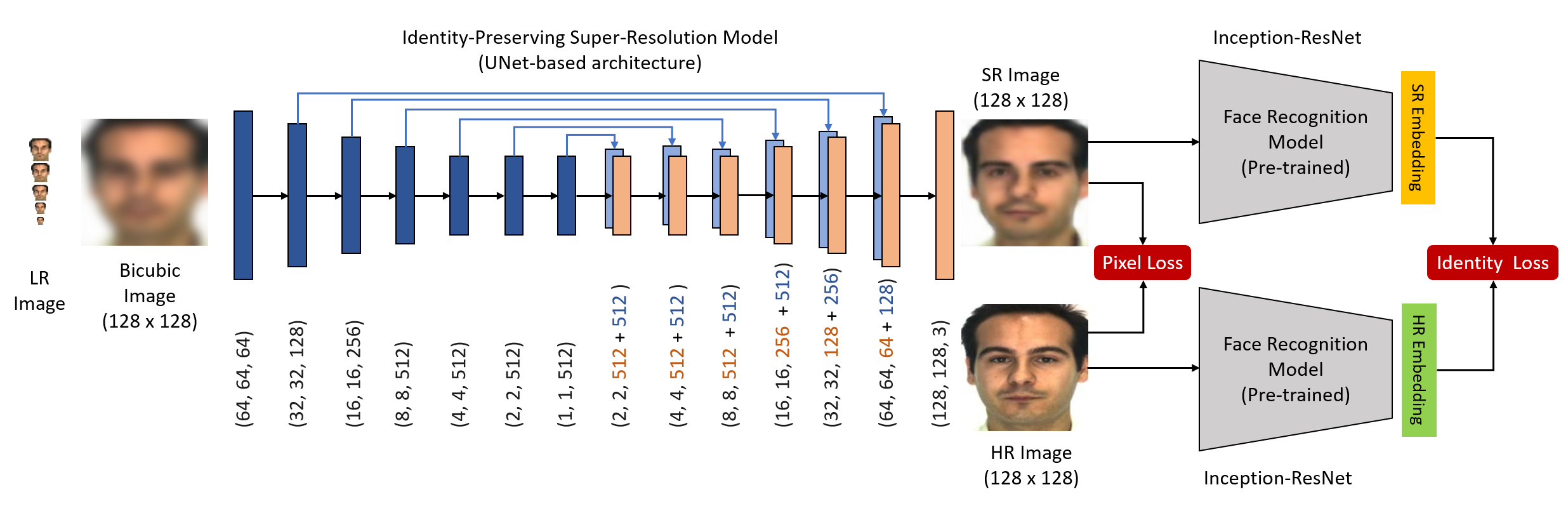}}
\caption{Model Architecture and training procedure. Our identity-preserveing encoder-decoder network consists of 7 downsampling and 7 upsampling blocks with symmetric contracting paths (skip connections). The encoder-decoder network is trained to translate the bicubic upsampled low-resolution faces (from various scales) to the super-resolved version (i.e. image-to-image translation). In training, we used a weighted loss which combines the pixel loss between input and output of the encoder-decoder network, and the identity loss measuring the distance between high resolution (HR) and super-resolved (SR) images in embedding space of a pretrained face recognition model. In testing, the low resolution (LR) face is super-resolved to SR and its embedding will be compared with embeddings of HR gallery images with a cosine distance measure.}\label{model_architecture} \centering
\end{figure*}

Figure \ref{model_architecture} depicts the proposed model architecture. It consists of an identity-preserving face hallucination image-to-image translation model and a pre-trained face recognition network. The deep convolutional face hallucination model translates the the bicubic upsampled low-resolution faces to super-resolved version during training (i.e. image-to-image translation). The pre-trained face recognition model produces discriminative, low-dimensional embeddings for both super-resolved and high-resolution faces in order to compute the identity loss.

\subsection{Loss Function}
\label{sssec:lossfunction}
In this section, we designed two loss objectives for this problem. Firstly, we included a loss to match the faces in the pixel level. In this loss, we calculated the differences between the super-resolved and high-resolution faces with mean squared error (MSE) loss defined in Equation \ref{pixel_loss}. Further, to achieve identity-preserving face hallucination at the same time, we used an identity-loss which is the MSE distance between the low-dimensional embeddings of the super-resolved and high-resolution faces extracted from the pre-trained face recognition model shown in Equation \ref{embedding_loss}. In our experiments, the second loss showed to be very helpful and increased the performance in comparison to the case when only the pixel loss is applied.

The pre-trained face recognition network \cite{schroff2015facenet} we utilized for our approach, is able to extract discriminative features from faces. These features are pose invariant and robust against different illumination conditions. Further, the data set that the model is trained on includes faces of various races.

\begin{center}
\begin{equation}
    L_{pixel} = \frac{1}{n} \sum_{i=1}^{n} \|F_i^{SR} - F_i^{HR}\|
     \label{pixel_loss}
\end{equation}
\end{center}
\begin{center}
\begin{equation}
    L_{embedding} = \frac{1}{n} \sum_{i=1}^{n} \|E_i^{SR} - E_i^{HR}\|
     \label{embedding_loss}
\end{equation}
\end{center}

We defined our total loss as a weighted sum of pixel and identity losses as described in Equation \ref{total_loss}:

\begin{center}
\begin{equation}
    L_{total} = L_{embedding} + \beta L_{pixel}
     \label{total_loss}
\end{equation}
\end{center}
We tried different weights for the two losses and after some experiments, we set $\beta$ to 5. The performance is not sensitive this hyperparameter.

\subsection{Training Procedure}
\label{sec:typestyle}

We trained our convolutional encoder-decoder deep network on the VGGface2 data set \cite{Cao2018} which includes more than 3.3 million faces and used a pre-trained face recognition on the MS-Celeb data set \cite{Guo2016}. In our setting, we employed the bicubic interpolation to resize the input of both networks. We aligned the VGGface2 data set using MTCNN \cite{zhang2016joint} and resized the images to 128x128 pixels before training. Some faces were deleted in the preprocessing step as they were not detected by MTCNN. After this preprocessing, we ended up with approximately 2.7 million faces.

During the training, firstly the faces were randomly downsampled to one of these resolutions: 7x6, 11x8, 14x12, 16x12, 16x14, 16x16, 18x16, 21x15, 32x32 and 112x96. Then, they were resized (with bicubic interpolation) to 128x128 which was the input size of our identity-preserving super-resolution network. The network mapped that bicubic version to a super-resolved face. By including multiple resolutions during the training, we intended to make the model robust to various real-world low-resolution conditions. In the training, the probability of each resolution being chosen was equal.

\section{Experiments and Results}
\label{sec:ex&res}

After training, we picked out three different data sets for testing the performance of our model in low-resolution face recognition tasks. The first data set was AR \cite{Martinez1998} which is a controlled data set and the faces are not low-resolution by their nature. The second data set we chose was LFW \cite{huang2008labeled} which is more challenging and has a lot of variations. SCface \cite{Grgic2011} was the third data set that we conducted our experiments on. The faces in this data set are captured from surveillance cameras with various distances and qualities. The distribution of this data set differs from our training set and the other test set. The images in this data set are naturally low-resolution.

\subsection{Low-Resolution Face Identification on a Controlled Face Data Set}
\label{sssec:ar_ex}

\begin{figure*}[t]
\centering
\includegraphics[scale=0.38]{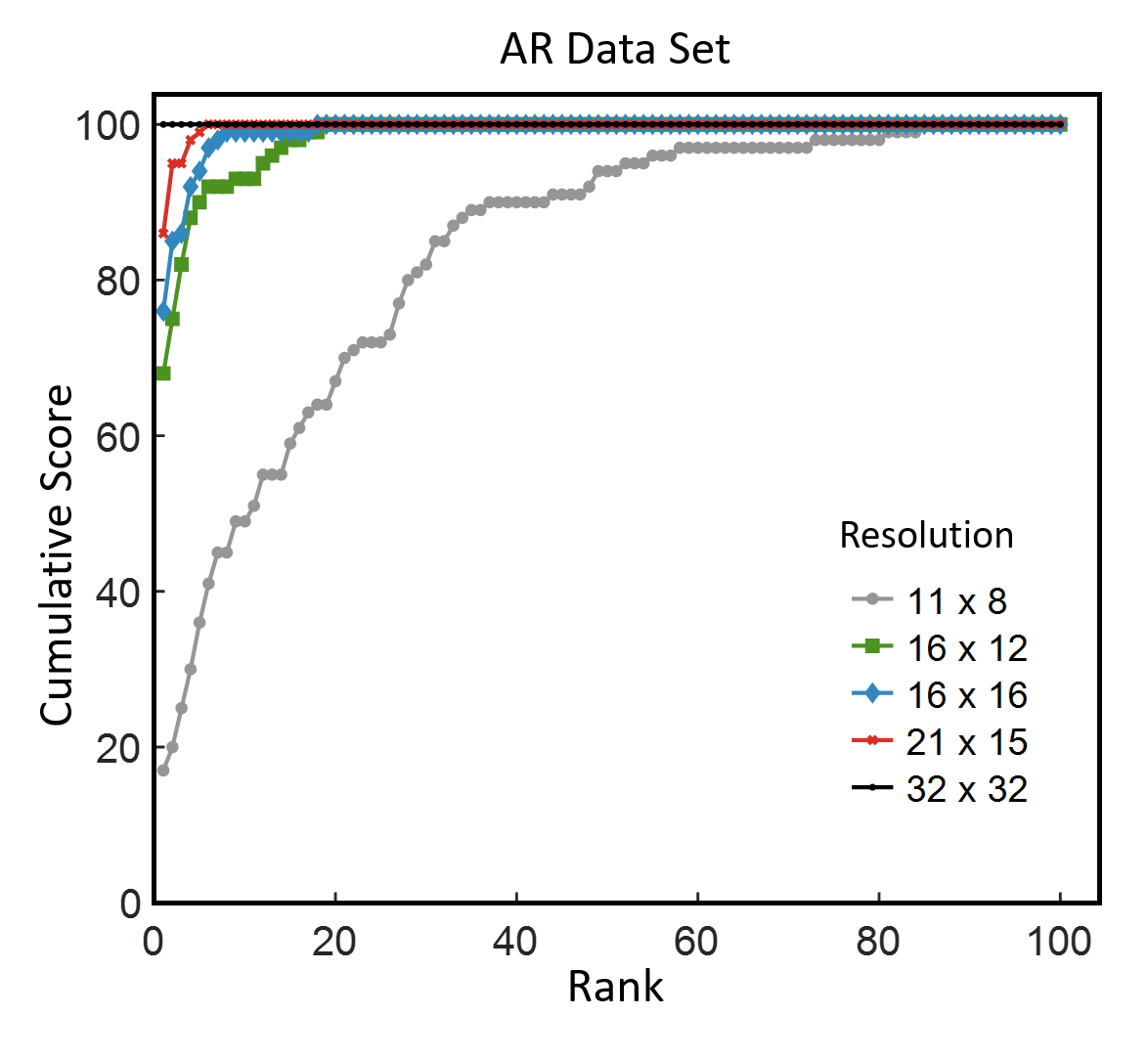}
\caption{Cumulative match characteristic for AR data set.}\label{AR_cmc}
\centering
\end{figure*}

The images of the AR data set were captured in different illumination conditions, facial expression and occlusion. It contains over 4,000 color images corresponding to 126 people's faces (70 men and 56 women). The reason we selected this data set was to evaluate our method on a controlled benchmark data set \cite{zhang2021dissimilarity}\cite{kar2020triangular}\cite{prasad2020deep}\cite{li2019low}. The faces are in a frontal view and under strictly controlled conditions. In this experiment following the setup in \cite{li2019low}, we selected 100 subjects (50 men, 50 women). For each subject one image from day 14 as gallery and one image from day 1 as probe. The gallery was resized to 128x128 and the probe was downsampled to 11x8, 16x12, 16x16, 21x15 and 32x32. We calculated the cumulative match characteristic up to rank-100 between the gallery and the super-resolved version of the probe. This shows whether there is the correct target identity within the first $n$ predicted identities. $n$ is called the rank and the cumulative match characteristic is the sum of these ranks. We emphasize that AR data set was not used for training and our model is only evaluated on this data set. As depicted in Figure \ref{AR_cmc}, our model performs very well for all low resolutions. 

\begin{table}[ht]
\centering
\caption{Comparison of Rank-1 correct match score on AR data set.}
\tabcolsep=0.08cm
 \begin{tabular}{||c c c c c c||} 
 \hline
 Method & 11x8 & 16x12 & 16x16 & 21x15 & 32x32 \\ [0.5ex] 

 \hline\hline
 Bicubic & 3\% & 16\% & - & 59\% & - \\ 
 SRGAN \cite{Ledig2017} & 3\% & 8\% & - & 30\% & - \\ 
 LapSRN \cite{Lai2019} & 4\% & 14\% & - & 37\% & - \\ 
 ScSR \cite{Yang2010} & 3\% & 23\% & - & 37\% & - \\ 
 \textbf{Ours} & \textbf{17\%} & \textbf{68\%} & \textbf{76\%} & \textbf{86\%} & \textbf{100\%} \\ [1ex] 
 \hline
 \end{tabular}
\label{AR_rank1}
\end{table}

\begin{figure*}[t]
\centerline{\includegraphics[scale=0.4]{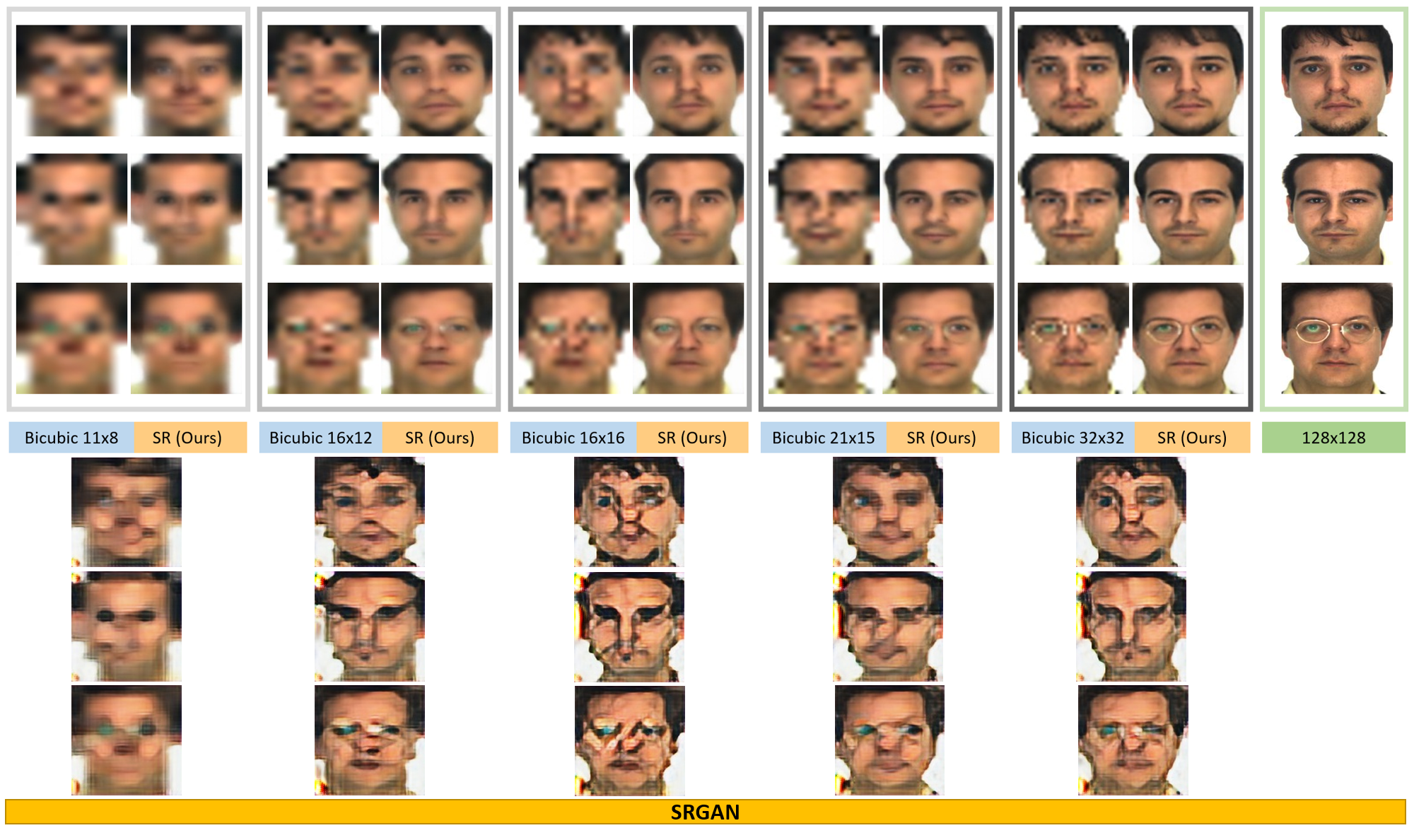}}
\caption{A visual demonstration of our proposed model performance in super resolving faces of AR data set for various low-resolution conditions.}\label{AR_vis_res}
\centering
\end{figure*}

Table \ref{AR_rank1} shows a comparison of rank-1 cumulative match score between our model and other state-of-the-art models on the AR data set. As demonstrated in Table \ref{AR_rank1}, in all tested low-resolution conditions, our model outperforms the competing methods significantly. This indicates that in controlled view condition and under variations of illumination, facial expression, and occlusion, our proposed model is the best performing model for low-resolution face recognition, and for resolutions 32x32 and above it is performing with 100 accuracy rate. Figure \ref{AR_vis_res} illustrates examples of the reconstructed faces in the AR data set with our proposed model for various input resolutions. As depicted, qualitatively, our proposed method also performs well in super-resolving low-resolution AR faces.

\begin{table}[ht]
\centering
\caption{Rank-1 identification rate (\%) on LFW}\label{lfw}
\tabcolsep=0.08cm
 \begin{tabular}{||c c c c c||} 
 \hline
Method & 7x6 & 14x12 & 16x14 & 18x16 \\ [0.5ex] 

 \hline\hline
SHI \cite{jian2015simultaneous} & - & \textbf{66.16\%} & 0.68.05\% & 0.69.2\%\\ 
Bicubic & 0.23\% & 8.04\% &	13.78\% &  23.01\% \\ 

EDSR\textsubscript{32,256,0 (VGG-Face)} \cite{lai2019low} & 5.13\% & 28.68\% & 32.59\% &  36.84\% \\
EDSR\textsubscript{32,256,0.5 (VGG-Face)} \cite{lai2019low} & 9.31\% & 39.32\% & 38.93\% &  43.26\% \\

EDSR\textsubscript{32,256,0 (SFace)} \cite{lai2019low} & 5.91\% & 50.78\% & 56.19\% &  65.89\% \\
EDSR\textsubscript{32,256,0.5 (SFace)} \cite{lai2019low} & \textbf{14.91\%} & 63.34\% & 63.78\% &  71.96\% \\
\textbf{Ours} & 3.16\% & 56.97\% & \textbf{69.77\%} & \textbf{78.72\%} \\ [1ex] 
 \hline
 \end{tabular}

\end{table}

\subsection{Low-Resolution Face Identification in the Wild}
To test the ability of our model in a more challenging and uncontrolled low-resolution face data set, we applied it to the LFW data set \cite{huang2008labeled} for a face identification task. The dataset contains more than 13,000 images of faces collected from the web. 1680 of the people pictured have two or more distinct photos in the dataset. For testing, we followed the protocol in \cite{lai2019low}. We applied MTCNN for aligning the faces and then we removed the identities with less than 4 images. We randomly selected one image as a gallery and the rest as probes and then we calculated the rank-1 identification rate. We did this setup 10 times and calculated the average of the results. The low resolutions we used in this experiment was 7x6, 14x12, 16x14 and 18x16 and the high resolution probe was 112x96. Table \ref{lfw} depicts the results of this experiment when compared with competing methods. As demonstrated in the table, our model shows performance improvement on this data set in 16x14 and 18x16 resolutions.

\subsection{Low-Resolution Face Identification on Images Captured by Surveillance Cameras}
\label{sssec:scface_ex}
 
SCface is a database of static images of human faces. Images were taken in uncontrolled indoor environment using five video surveillance cameras of various qualities. Database contains 4160 static images (in visible and infrared spectrum) of 130 identities. Since there is no high-resolution pair for natural low-resolution faces, we did not train or fine-tune our model on this data set and only evaluated our model on it. We carried out two sets of experiments with two different protocols as defined in \cite{yang2017discriminative}. In the first testing setup, we designated the high-resolution frontal faces as galleries and the other faces from five different cameras and 3 various distances (1 meter, 2.6 meters and 4.7 meters) as probes. In the second setup, we have selected 50 images for each identity from the nearest camera (1 meter) as galleries and the rest from the 2.6m distance as probes. We resized the images to 64x64 and then upsampled them to 128x128 with bicubic interpolation to feed them to the model. We concatenated the embeddings of both upsampled face and super-resolved version to increase the performance.

\begin{table}[ht]
\centering
\caption{Rank-1 identification rate (\%) on SCface data set when the high-resolution frontal faces are used as galleries and the other faces from three standoff distances (1 meter, 2.6 meters and 4.7 meters) as probes.} 
\tabcolsep=0.08cm
 \begin{tabular}{||c c c c||} 
 \hline
 Method & 1m Camera & 2.6m Camera & 4.7m Camera \\ [0.5ex] 

 \hline\hline
 SCface \cite{Grgic2011} & 6.18\% & 6.18\% & 1.82\% \\ 
 CLPM \cite{li2009low} & 3.08\% & 4.32\% & 3.46\% \\
 SSR \cite{yang2014salient} & 18.09\% & 13.2\% & 7.04\% \\
 CSCDN \cite{wang2015low} & 18.97\% & 13.58\% & 6.99\% \\
 CCA \cite{wen2016discriminative} & 20.69\% & 14.85\% & 9.79\% \\
 DCA \cite{Haghighat2017} & 25.53\% & 18.44\% & 12.19\% \\
 C-RSDA \cite{chu2017low} & 18.46\% & 18.08\% & 15.77\% \\
 Centerloss \cite{Wen2016} & 31.71\% & 20.80\% & \textbf{20.40\%} \\
 LMSoftmax \cite{li2019low}& 18.00\% & 16.00\% & 14.00\% \\
 AMSoftmax \cite{li2019low} & 18.40\% & 20.80\% & 14.80\% \\ 
 L2softmax \cite{li2019low} & 16.80\% & 18.80\% & 9.20\% \\

 \textbf{Ours} & \textbf{96.44\%} & \textbf{74.10\%} & 9.70\%  \\ [1ex] 
 \hline
 \end{tabular}
\label{Rank1_scface_ex1}
\end{table}

\begin{table}[ht]
\centering
\caption{Rank-1 identification rate (\%) on SCface data set when faces from 1m standoff distance are used as galleries and faces from 2.6m standoff distance as probes.}\label{Rank1_scface_ex2}
\tabcolsep=0.08cm
 \begin{tabular}{||c c||} 
 \hline
 Method & 1m - 2.6m Camera \\ [0.5ex] 

 \hline\hline
 CLPM \cite{li2009low} & 29.12\% \\
 CMFA \cite{siena2012coupled} & 39.56\% \\
 LMsoftmax \cite{liu2016large} & 39.56\% \\ 
 SDA \cite{zhu2012face} & 40.08\% \\
 LMSoftmax \cite{li2019low} & 40.4\%  \\ 
 L2softmax \cite{ranjan2017l2} & 42.8\%  \\ 
 Coupled mapping method \cite{shi2014local} & 43.24\% \\
 AMSoftmax \cite{li2019low} & 46.8\%  \\ 
 LMCM \cite{zhang2016learning} & 60.40\% \\
 Centerloss \cite{li2019low} & 69.60\%  \\ 
 \textbf{Ours} & \textbf{76.89\%}  \\ [1ex]
 \hline
 \end{tabular}
\end{table}

As depicted in Table \ref{Rank1_scface_ex1} and \ref{Rank1_scface_ex2}, our model is the best performing model. Considering the fact that our model was not trained on this data set, the demonstrated performance shows the considerable generalizability of our proposed model. This is especially critical in real-world low-resolution face recognition applications.

\section{Conclusion}
\label{sec:conclusion}

In this paper, we proposed a deep convolutional identity-preserving encoder-decoder network with skip paths to perform better face hallucination for low-resolution face recognition. In the training process, we incorporated a combination of pixel level and identity loss to achieve this objective. This showed that the identity loss is very helpful and increased the performance especially in very low-resolution cases. To gain robustness against different real-world low-resolution conditions, we trained the model in a multi-scale manner. In our experiments, we tested the model on both natural and artificial low-resolution faces. Our results showed that our identity-preserving model outperforms state-of-the-art models in low-resolution face recognition task in various experimental settings. However, there might be some cases that the model fails in recovering identity when the faces have many variations like high orientations, make-up, occlusion etc. For our future work, we are going to benefit from face landmarks to localize the face features to solve this issue.

Furthermore, we are working on a mechanism to upsample the input of the models without using any interpolation. We believe these enhancements will increase the performance even more. Furthermore, the proposed model demonstrated considerable generalizability to challenging data sets which were not used in training. This feature is especially important in real-world applications of low-resolution face recognition systems.

\printbibliography[heading=subbibintoc]

@inproceedings{huang2008labeled,
    title = {Labeled Faces in the Wild: A Database for Studying Face Recognition in Unconstrained Environments},
  author={Huang, Gary B and Mattar, Marwan and Berg, Tamara and Learned-Miller, Eric},
  doi = {10.1117/12.2080393},
  booktitle = {Technical Report},
  year={2008}
}

@article{Chen2020,
author = {Chen, Jin and Chen, Jun and Wang, Zheng and Liang, Chao and Lin, Chia Wen},
doi = {10.1109/LSP.2020.2986942},
issn = {15582361},
journal = {IEEE Signal Processing Letters},
number = {c},
pages = {645--649},
title = {{Identity-Aware Face Super-Resolution for Low-Resolution Face Recognition}},
volume = {27},
year = {2020}
}

@article{Dong2016,
archivePrefix = {arXiv},
arxivId = {1501.00092},
author = {Dong, Chao and Loy, Chen Change and He, Kaiming and Tang, Xiaoou},
doi = {10.1109/TPAMI.2015.2439281},
eprint = {1501.00092},
issn = {01628828},
journal = {IEEE Transactions on Pattern Analysis and Machine Intelligence},
number = {2},
pages = {295--307},
pmid = {26761735},
title = {{Image Super-Resolution Using Deep Convolutional Networks}},
volume = {38},
year = {2016}
}

@article{Hsu2019,
archivePrefix = {arXiv},
arxivId = {1807.08370},
author = {Hsu, Chih Chung and Lin, Chia Wen and Su, Weng Tai and Cheung, Gene},
doi = {10.1109/TIP.2019.2924554},
eprint = {1807.08370},
issn = {19410042},
journal = {IEEE Transactions on Image Processing},
number = {12},
pages = {6225--6236},
pmid = {31265397},
title = {{SiGAN: Siamese Generative Adversarial Network for Identity-Preserving Face Hallucination}},
volume = {28},
year = {2019}
}

@article{Yu2020,
author = {Yu, Xin and Porikli, Fatih and Fernando, Basura and Hartley, Richard},
doi = {10.1007/s11263-019-01254-5},
issn = {15731405},
journal = {International Journal of Computer Vision},
number = {2},
pages = {500--526},
title = {{Hallucinating Unaligned Face Images by Multiscale Transformative Discriminative Networks}},
volume = {128},
year = {2020}
}

@article{Martinez1998,
author = {Martinez, Aleix and Benavente, Robert},
doi = {10.1023/B:VISI.0000029666.37597},
isbn = {24},
journal = {CVC Technical Report},
pages = {1--10},
title = {{The AR face database}},
volume = {24},
year = {1998}
}

@article{Lai2019,
archivePrefix = {arXiv},
arxivId = {1710.01992},
author = {Lai, Wei Sheng and Huang, Jia Bin and Ahuja, Narendra and Yang, Ming Hsuan},
doi = {10.1109/TPAMI.2018.2865304},
eprint = {1710.01992},
issn = {19393539},
journal = {IEEE Transactions on Pattern Analysis and Machine Intelligence},
number = {11},
pages = {2599--2613},
pmid = {30106708},
title = {{Fast and Accurate Image Super-Resolution with Deep Laplacian Pyramid Networks}},
volume = {41},
year = {2019}
}

@article{Yang2010,
author = {Yang, Jianchao and Wright, John and Huang, Thomas S. and Ma, Yi},
doi = {10.1109/TIP.2010.2050625},
issn = {10577149},
journal = {IEEE Transactions on Image Processing},
number = {11},
pages = {2861--2873},
title = {{Image super-resolution via sparse representation}},
volume = {19},
year = {2010}
}

@article{zangeneh2020low,
  title={Low resolution face recognition using a two-branch deep convolutional neural network architecture},
  author={Zangeneh, Erfan and Rahmati, Mohammad and Mohsenzadeh, Yalda},
  journal={Expert Systems with Applications},
  volume={139},
  pages={112854},
  year={2020},
  publisher={Elsevier}
}

@article{He2019,
author = {He, Chenhang and Lai, Shuncheung and Lam, Kinman},
isbn = {9781538646588},
journal = {2019 IEEE International Conference on Image Processing (ICIP)},
pages = {2537--2541},
publisher = {IEEE},
title = {{LOW-RESOLUTION FACE RECOGNITION BASED ON IDENTITY-PRESERVED FACE HALLUCINATION}},
year = {2019}
}

@article{Ledig2017,
archivePrefix = {arXiv},
arxivId = {arXiv:1609.04802v5},
author = {Ledig, Christian and Theis, Lucas and Husz{\'{a}}r, Ferenc and Caballero, Jose and Cunningham, Andrew and Acosta, Alejandro and Aitken, Andrew and Tejani, Alykhan and Totz, Johannes and Wang, Zehan and Shi, Wenzhe},
doi = {10.1109/CVPR.2017.19},
eprint = {arXiv:1609.04802v5},
isbn = {9781538604571},
journal = {Proceedings - 30th IEEE Conference on Computer Vision and Pattern Recognition, CVPR 2017},
pages = {105--114},
title = {{Photo-realistic single image super-resolution using a generative adversarial network}},
volume = {2017-Janua},
year = {2017}
}

@article{Yu2017,
author = {Yu, Xin and Porikli, Fatih},
journal = {31st AAAI Conference on Artificial Intelligence, AAAI 2017},
pages = {4327--4333},
title = {{Face hallucination with tiny unaligned images by transformative discriminative neural networks}},
year = {2017}
}

@inproceedings{zhang2018low,
  title={Low resolution face recognition and reconstruction via deep canonical correlation analysis},
  author={Zhang, Zhao and Yuan, Yun-Hao and Shen, Xiao-Bo and Li, Yun},
  booktitle={2018 IEEE International Conference on Acoustics, Speech and Signal Processing (ICASSP)},
  pages={2951--2955},
  year={2018},
  organization={IEEE}
}

@inproceedings{noor2019gradient,
  title={Gradient image super-resolution for low-resolution image recognition},
  author={Noor, Dewan Fahim and Li, Yue and Li, Zhu and Bhattacharyya, Shuvra and York, George},
  booktitle={ICASSP 2019-2019 IEEE International Conference on Acoustics, Speech and Signal Processing (ICASSP)},
  pages={2332--2336},
  year={2019},
  organization={IEEE}
}

@inproceedings{zhu2019low,
  title={Low-resolution Visual Recognition via Deep Feature Distillation},
  author={Zhu, Mingjian and Han, Kai and Zhang, Chao and Lin, Jinlong and Wang, Yunhe},
  booktitle={ICASSP 2019-2019 IEEE International Conference on Acoustics, Speech and Signal Processing (ICASSP)},
  pages={3762--3766},
  year={2019},
  organization={IEEE}
}

@inproceedings{li2018face,
  title={Face hallucination based on key parts enhancement},
  author={Li, Ke and Bare, Bahetiyaer and Yan, Bo and Feng, Bailan and Yao, Chunfeng},
  booktitle={2018 IEEE International Conference on Acoustics, Speech and Signal Processing (ICASSP)},
  pages={1378--1382},
  year={2018},
  organization={IEEE}
}

@article{Ronneberger2015,
archivePrefix = {arXiv},
arxivId = {arXiv:1505.04597v1},
author = {Ronneberger, Olaf and Fischer, Philipp and Brox, Thomas},
doi = {10.1007/978-3-319-24574-4_28},
eprint = {arXiv:1505.04597v1},
isbn = {9783319245737},
issn = {16113349},
journal = {Lecture Notes in Computer Science},
pages = {234--241},
title = {{U-net: Convolutional networks for biomedical image segmentation}},
volume = {9351},
year = {2015}
}

@article{Schroff2015,
archivePrefix = {arXiv},
arxivId = {1503.03832},
author = {Schroff, Florian and Kalenichenko, Dmitry and Philbin, James},
doi = {10.1109/CVPR.2015.7298682},
eprint = {1503.03832},
isbn = {9781467369640},
issn = {10636919},
journal = {Proceedings of the IEEE Computer Society Conference on Computer Vision and Pattern Recognition},
pages = {815--823},
title = {{FaceNet: A unified embedding for face recognition and clustering}},
volume = {07-12-June},
year = {2015}
}

@article{Cao2018,
archivePrefix = {arXiv},
arxivId = {1710.08092},
author = {Cao, Qiong and Shen, Li and Xie, Weidi and Parkhi, Omkar M. and Zisserman, Andrew},
doi = {10.1109/FG.2018.00020},
eprint = {1710.08092},
isbn = {9781538623350},
journal = {Proceedings - 13th IEEE International Conference on Automatic Face and Gesture Recognition, FG 2018},
pages = {67--74},
title = {{VGGFace2: A dataset for recognising faces across pose and age}},
year = {2018}
}

@article{Guo2016,
archivePrefix = {arXiv},
arxivId = {arXiv:1607.08221v1},
author = {Guo, Yandong and Zhang, Lei and Hu, Yuxiao and He, Xiaodong and Gao, Jianfeng},
doi = {10.1007/978-3-319-46487-9_6},
eprint = {arXiv:1607.08221v1},
isbn = {9783319464862},
issn = {16113349},
journal = {Lecture Notes in Computer Science},
pages = {87--102},
title = {{MS-celeb-1M: A dataset and benchmark for large-scale face recognition}},
volume = {9907 LNCS},
year = {2016}
}

@article{zhang2016joint,
  title={Joint face detection and alignment using multitask cascaded convolutional networks},
  author={Zhang, Kaipeng and Zhang, Zhanpeng and Li, Zhifeng and Qiao, Yu},
  journal={IEEE Signal Processing Letters},
  volume={23},
  number={10},
  pages={1499--1503},
  year={2016},
  publisher={IEEE}
}

@article{Grgic2011,
author = {Grgic, Mislav and Delac, Kresimir and Grgic, Sonja},
doi = {10.1007/s11042-009-0417-2},
issn = {13807501},
journal = {Multimedia Tools and Applications},
number = {3},
pages = {863--879},
title = {{SCface - Surveillance cameras face database}},
volume = {51},
year = {2011}
}

@article{Haghighat2017,
author = {Haghighat, Mohammad and Abdel-Mottaleb, Mohamed},
doi = {10.1109/FG.2017.130},
isbn = {9781509040230},
journal = {Proceedings - 12th IEEE International Conference on Automatic Face and Gesture Recognition},
pages = {912--917},
title = {{Lower Resolution Face Recognition in Surveillance Systems Using Discriminant Correlation Analysis}},
year = {2017}
}

@inproceedings{liu2016large,
  title={Large-margin softmax loss for convolutional neural networks.},
  author={Liu, Weiyang and Wen, Yandong and Yu, Zhiding and Yang, Meng},
  booktitle={ICML},
  volume={2},
  pages={7},
  year={2016}
}

@article{Wen2016,
archivePrefix = {arXiv},
arxivId = {1311.2901},
author = {Wen, Yandong and Zhang, Kaipeng and B, Zhifeng Li and Qiao, Yu},
doi = {10.1007/978-3-319-46478-7},
eprint = {1311.2901},
isbn = {9783319464787},
issn = {0302-9743},
journal = {Eccv},
pages = {499--515},
pmid = {4520227},
title = {{A Discriminative Feature Learning Approach}},
volume = {1},
year = {2016}
}

@article{ranjan2017l2,
  title={L2-constrained softmax loss for discriminative face verification},
  author={Ranjan, Rajeev and Castillo, Carlos D and Chellappa, Rama},
  journal={arXiv preprint arXiv:1703.09507},
  year={2017}
}

@article{yang2017discriminative,
  title={Discriminative multidimensional scaling for low-resolution face recognition},
  author={Yang, Fuwei and Yang, Wenming and Gao, Riqiang and Liao, Qingmin},
  journal={IEEE Signal Processing Letters},
  volume={25},
  number={3},
  pages={388--392},
  year={2017},
  publisher={IEEE}
}

@article{li2019low,
  title={On low-resolution face recognition in the wild: Comparisons and new techniques},
  author={Li, Pei and Prieto, Loreto and Mery, Domingo and Flynn, Patrick J},
  journal={IEEE Transactions on Information Forensics and Security},
  volume={14},
  number={8},
  pages={2000--2012},
  year={2019},
  publisher={IEEE}
}

@inproceedings{schroff2015facenet,
  title={Facenet: A unified embedding for face recognition and clustering},
  author={Schroff, Florian and Kalenichenko, Dmitry and Philbin, James},
  booktitle={Proceedings of the IEEE conference on computer vision and pattern recognition},
  pages={815--823},
  year={2015}
}

@article{kar2020triangular,
  title={Triangular coil pattern of local radius of gyration face for heterogeneous face recognition},
  author={Kar, Arindam and Neogi, Pinaki Prasad Guha},
  journal={Applied Intelligence},
  volume={50},
  number={3},
  pages={698--716},
  year={2020},
  publisher={Springer}
}

@incollection{prasad2020deep,
  title={Deep learning based representation for face recognition},
  author={Prasad, Puja S and Pathak, Rashmi and Gunjan, Vinit Kumar and Rao, HV Ramana},
  booktitle={ICCCE 2019},
  pages={419--424},
  year={2020},
  publisher={Springer}
}

@article{zhang2021dissimilarity,
  title={Dissimilarity-based nearest neighbor classifier for single-sample face recognition},
  author={Zhang, Zhengqi and Zhang, Li and Zhang, Meng},
  journal={The Visual Computer},
  volume={37},
  number={4},
  pages={673--684},
  year={2021},
  publisher={Springer}
}

@inproceedings{lai2019low,
  title={Low-resolution face recognition based on identity-preserved face hallucination},
  author={Lai, Shun-Cheung and He, Chen-Hang and Lam, Kin-Man},
  booktitle={2019 IEEE International Conference on Image Processing (ICIP)},
  pages={1173--1177},
  year={2019},
  organization={IEEE}
}

@article{jian2015simultaneous,
  title={Simultaneous hallucination and recognition of low-resolution faces based on singular value decomposition},
  author={Jian, Muwei and Lam, Kin-Man},
  journal={IEEE Transactions on Circuits and Systems for Video Technology},
  volume={25},
  number={11},
  pages={1761--1772},
  year={2015},
  publisher={IEEE}
}

@article{li2009low,
  title={Low-resolution face recognition via coupled locality preserving mappings},
  author={Li, Bo and Chang, Hong and Shan, Shiguang and Chen, Xilin},
  journal={IEEE Signal processing letters},
  volume={17},
  number={1},
  pages={20--23},
  year={2009},
  publisher={IEEE}
}

@inproceedings{siena2012coupled,
  title={Coupled marginal fisher analysis for low-resolution face recognition},
  author={Siena, Stephen and Boddeti, Vishnu Naresh and Vijaya Kumar, BVK},
  booktitle={European Conference on Computer Vision},
  pages={240--249},
  year={2012},
  organization={Springer}
}

@inproceedings{yang2014salient,
  title={Salient color names for person re-identification},
  author={Yang, Yang and Yang, Jimei and Yan, Junjie and Liao, Shengcai and Yi, Dong and Li, Stan Z},
  booktitle={European conference on computer vision},
  pages={536--551},
  year={2014},
  organization={Springer}
}

@article{wang2015low,
  title={Low-resolution degradation face recognition over long distance based on CCA},
  author={Wang, Zhenyu and Yang, Wankou and Ben, Xianye},
  journal={Neural Computing and Applications},
  volume={26},
  number={7},
  pages={1645--1652},
  year={2015},
  publisher={Springer}
}

@inproceedings{wen2016discriminative,
  title={A discriminative feature learning approach for deep face recognition},
  author={Wen, Yandong and Zhang, Kaipeng and Li, Zhifeng and Qiao, Yu},
  booktitle={European conference on computer vision},
  pages={499--515},
  year={2016},
  organization={Springer}
}

@article{chu2017low,
  title={Low-resolution face recognition with single sample per person},
  author={Chu, Yongjie and Ahmad, Touqeer and Bebis, George and Zhao, Lindu},
  journal={Signal Processing},
  volume={141},
  pages={144--157},
  year={2017},
  publisher={Elsevier}
}

@inproceedings{zhu2012face,
  title={Face detection, pose estimation, and landmark localization in the wild},
  author={Zhu, Xiangxin and Ramanan, Deva},
  booktitle={2012 IEEE conference on computer vision and pattern recognition},
  pages={2879--2886},
  year={2012},
  organization={IEEE}
}

@inproceedings{zhang2016learning,
  title={Learning a discriminative null space for person re-identification},
  author={Zhang, Li and Xiang, Tao and Gong, Shaogang},
  booktitle={Proceedings of the IEEE conference on computer vision and pattern recognition},
  pages={1239--1248},
  year={2016}
}

@article{shi2014local,
  title={From local geometry to global structure: Learning latent subspace for low-resolution face image recognition},
  author={Shi, Jingang and Qi, Chun},
  journal={IEEE Signal Processing Letters},
  volume={22},
  number={5},
  pages={554--558},
  year={2014},
  publisher={IEEE}
}

\end{document}